\documentclass[conference]{IEEEtran}
\IEEEoverridecommandlockouts
\usepackage{cite}
\usepackage{amsmath,amssymb,amsfonts}
\usepackage{algorithmic}
\usepackage{graphicx}
\usepackage{textcomp}
\usepackage{xcolor}
\usepackage{float}
\usepackage{svg}
\usepackage[inline]{enumitem}
\usepackage{booktabs}
\usepackage{tabularx}
\usepackage[utf8]{inputenc}
\usepackage{tikz}
\usetikzlibrary{calc}

\def\BibTeX{{\rm B\kern-.05em{\sc i\kern-.025em b}\kern-.08em
    T\kern-.1667em\lower.7ex\hbox{E}\kern-.125emX}}

\usepackage[
  top=0.75in,
  bottom=1.03in,
  left=0.625in,
  right=0.625in
]{geometry}

\begin{document}

\title{OsteoCAD: A Human-in-the-Loop Cloud–Edge Framework for Bone Tumor Segmentation}

\author{\IEEEauthorblockN{Maximo Rodriguez-Herrero\IEEEauthorrefmark{1}, Dante D. Sanchez-Gallegos\IEEEauthorrefmark{1}, Heriberto Aguirre-Meneses\IEEEauthorrefmark{2}, \\ Marco Antonio N\'u\~{n}ez-Gaona\IEEEauthorrefmark{2}, J. L. Gonzalez-Compean\IEEEauthorrefmark{3}, and Jesus Carretero\IEEEauthorrefmark{1}}
\IEEEauthorblockA{\IEEEauthorrefmark{1}Department of Computer Science\\
University Carlos III of Madrid, Leganes, Spain\\
Email: maxrodri@inf.uc3m.es, dantsanc@pa.uc3m.es}
\IEEEauthorblockA{\IEEEauthorrefmark{2}Instituto Nacional de Rehabilitacion ``Luis Guillermo Ibarra Ibarra'', Mexico City, Mexico}
\IEEEauthorblockA{\IEEEauthorrefmark{3}Cinvestav Tamaulipas, Cd. Victoria, Mexico}}

\maketitle

\begin{abstract}
Artificial Intelligence (AI) and Deep Learning (DL) have notably advanced medical image analysis, yet many healthcare organizations struggle to adopt them due to limited computational resources and specialized expertise. To address these barriers, we introduce \textbf{OsteoCAD}, a modular eHealth framework that democratizes access to DL tools in clinical practice. OsteoCAD delivers end-to-end DL capabilities---from dataset creation and preprocessing to model training and inference---through an integrated and user-friendly interface. To mitigate local hardware constraints, the framework securely connects to remote GPU infrastructures. We validate OsteoCAD’s feasibility through a real-world case study in Mexico focused on large bone tumor segmentation. The results demonstrate the framework’s ability to enable DL-powered eHealth solutions without demanding advanced technical expertise or complex local configurations.
\end{abstract}


\section{Introduction}

In recent years, advances in Artificial Intelligence (AI) have greatly enhanced the automation of various processes within eHealth systems. The primary applications of AI in medical imaging assist physicians and specialists in identifying regions of interest, thereby reducing the time required for diagnosis and prognosis~\cite{Litjens2017AAnalysis, Sahiner2019DeepTherapy}. Consequently, AI-driven systems hold the potential to enable adaptive and personalized treatments, real-time monitoring, and improved diagnostic support~\cite{Fujita2020AI-basedFirst}.

Deep learning-based image segmentation, led by architectures such as U-Net and its variants~\cite{Ronneberger2015U-net:Segmentation,IsenseeNnU-NetSegmentation,Siddique2021U-NetApplications}, has dominated major medical image segmentation benchmarks in recent years. Despite this success, clinical feasibility remains challenging due to scarce labeled data, strict data protection regulations that limit dataset sharing, and the technical complexity of configuring, training, and maintaining these models in hospital environments. These limitations raise a central question for this work: \textit{How feasible is it to translate cutting-edge research models into real-world hospital environments?} They also motivate our focus on a framework that supports human-in-the-loop dataset expansion, privacy-preserving data handling, and simplified deployment workflows.

Beyond model architecture, image segmentation in eHealth environments requires addressing several additional needs like \textit{i)} providing an intuitive data visualization mechanism, \textit{ii)} the creation of a reliable, secure, and private mechanism for sharing and managing data, \textit{iii)} efficient workflows for data  management, processing, and labeling, \textit{iv)} support model explainability and user trust, and \textit{v)} providing   accessible deployment and maintenance pipelines~\cite{Hassan2024BarriersReview.,Liu2025ImprovingReview}. Furthermore, the adoption of DL solutions remains largely limited to well-resourced institutions. Smaller or resource-constrained clinics often lack the computational infrastructure, data volume, or technical expertise required to develop and deploy these models effectively beyond pilot projects~\cite{MaimaitiailiArtificialAuthor, Kmietowicz2017RadiologistCollege}. While public clouds have democratized access to resources across industries and research areas, their use in healthcare settings may raise privacy concerns, complicating adoption.

Considering these challenges, we present \textbf{OsteoCAD}, an eHealth framework designed for resource-constrained clinical environments. OsteoCAD combines a modular, container-based architecture with secure, remote GPU execution, allowing hospitals to build, train, and deploy segmentation models without local high-performance hardware. Unlike general-purpose machine learning platforms, OsteoCAD targets clinical workflows end-to-end, from dataset creation and human-in-the-loop annotation to deployment of segmentation models within routine hospital practice.
By lowering technical and infrastructural barriers, OsteoCAD aims to facilitate access to computer-assisted diagnostics across medical centers, regardless of their computational resources. Moreover, OsteoCAD is agnostic to the target segmentation label, so it can be used by multiple teams with different segmentation objectives, whether they share the same patient dataset with different annotations or work on entirely separate datasets.

We validate the proposed framework through a real-world use case focused on the detection of large-bone tumors in computed tomography (CT) scans at a clinic with 
\textbf{no existing datasets} and \textbf{no prior AI deployment} 
in clinical routine. This evaluation encompasses (i) the incremental construction of a labeled dataset comprising \textbf{67 patients} approved under institutional research protocols, (ii) model performance assessment using standard medical imaging metrics, achieving a Dice score of \textbf{0.84~$\pm$~0.02} on an initial cohort of 16 patients, and (iii) demonstration of the framework’s ability to address a complete, real-world clinical deployment scenario.

In summary, the contributions of this paper are:
\begin{enumerate}
    \item The design and implementation of a modular, DL framework enabling remote GPU-powered segmentation workflows in resource-constrained hospitals.
    \item The clinical deployment for large-bone tumor segmentation with human-in-the-loop dataset expansion and reported performance in a real hospital setting.
\end{enumerate}

The rest of the paper is organized as follows: \S~\ref{sec:related-work} reviews related work. \S~\ref{sec:system-architecture} describes the OsteoCAD framework architecture. \S~\ref{sec:clinical-usecase} presents a real-world use case. Finally, \S~\ref{sec:conclusions} provides a brief discussion and conclusions.

\section{Background and Related Work}
\label{sec:related-work}

\textbf{Medical Imaging on eHealth Environments}. A rich ecosystem of tools exists across desktop, web, and integrated platforms. Desktop solutions such as 3D Slicer, MITK, and ITK-SNAP offer comprehensive visualization, manual/semi-automatic segmentation, and plugin extensibility but face deployment barriers including OS dependencies, installation overhead, and restricted physical access \cite{Fedorov20123DNetwork,Wolf2004TheITK,Yushkevich2016ITK-SNAP:Images}. 

Web-based viewers like OHIF (DICOMweb with segmentation overlays) and Niivue (WebGL NIfTI rendering) solve accessibility issues but lack integrated AI training pipelines \cite{ZieglerOpenResearch,Overbeek2021Niivue:Meshes}. Comprehensive platforms such as XNAT with NVIDIA Clara and MONAI Label provide end-to-end AI annotation, training, and deployment with human-in-the-loop capabilities and CVAT integration, but typically target larger institutions with dedicated IT infrastructure \cite{Herrick2015XNATData,Zhu2025Parabricks:Suite,Diaz-Pinto2024MONAIImages}.

Similar to these tools, OsteoCAD enables the visualization of data using a web-based, modular design. Despite other tools in the literature, OsteoCAD is flexible enough to adopt or integrate with more mature software stacks, such as MONAI Label+OHIF+XNAT, thereby enabling reconstruction of equivalent functionality with less custom development. These established platforms provide DICOM compliance, security features, and active communities that OsteoCAD could leverage to scale for production.


OsteoCAD does not aim to outperform specialized tools at their core competencies (e.g., 3D Slicer's visualization, MONAI Label's annotation strategies). Instead, it serves as a lightweight bridge---enabling resource-constrained clinics to test AI-assisted segmentation with minimal upfront investment. Starting from zero datasets, OsteoCAD delivers working models through a few human-in-the-loop iterations, proving value before committing to enterprise-scale platforms.


\textbf{Medical Image Segmentation}. OsteoCAD adopts the \textit{nnU-Net} framework~\cite{Isensee2018NnU-Net:Segmentation} as its core segmentation engine due to its proven robustness and minimal configuration overhead. \textit{nnU-Net} is a self-configuring system built upon the U-Net architecture that bridges the research-to-practice gap. It greatly simplifies training and tuning by automating key steps, such as data preprocessing, postprocessing, and ensembling. Its robustness has been validated through consistent top performance across numerous medical segmentation challenges. Yet, despite its remarkable automation and reproducibility, nnU-Net's practical deployment still requires suitable hardware, well-structured datasets, and technical expertise for configuration, deployment, and maintenance.

\section{System Architecture}
\label{sec:system-architecture}

OsteoCAD is designed as a modular, container-based framework for deploying deep learning workflows in healthcare environments. Its architecture targets two main constraints common in resource-limited institutions: scarce computational resources and limited access to AI expertise. To address these, OsteoCAD structures the workflow into independently deployable components that encapsulate data management, model training, and inference, while minimizing the need for advanced system configuration.

To this end, we designed OsteoCAD with a three-layer modular architecture based on virtual containers, enabling straightforward on-premises deployment and plug-and-play operation. Each layer can be deployed independently and communicates using standard protocols like HTTPS or SSH. This flexibility allows organizations to tailor deployments to their specific infrastructure. Specifically tailored for hospital environments, the system preserves local data sovereignty while optionally enabling access to remote GPU clusters through secure SSH tunneling. This hybrid design addresses common computational constraints in resource-limited settings, such as the use case of the Luis Guillermo Ibarra Ibarra National Rehabilitation Center (INR-LGII) described in Section \ref{sec:clinical-usecase}.

\begin{figure}
\centering
\includegraphics[width=\linewidth]{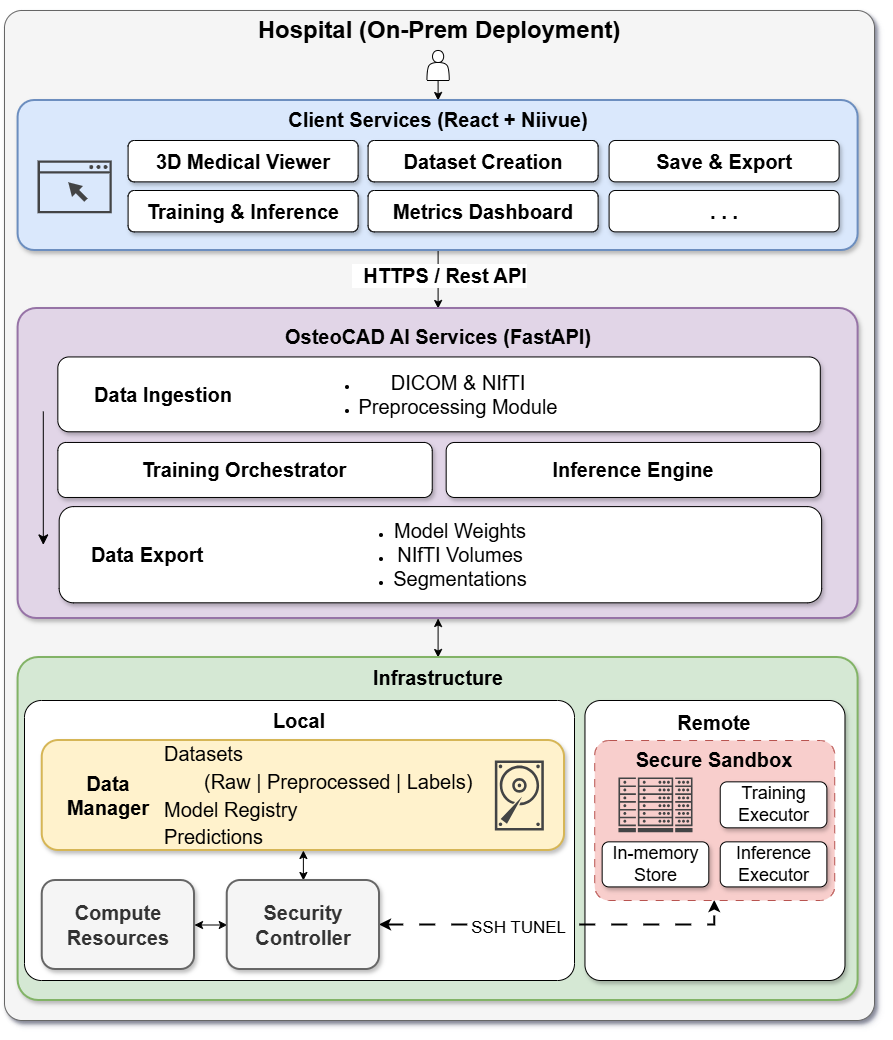}
\caption{Modular OsteoCAD architecture: FastAPI orchestrates local data processing and remote GPU execution while the React+Niivue frontend enables clinician interaction.}
\label{fig:osteocad-arch}
\end{figure}

\subsection{Architectural Layers}
\label{subsec:architectural-layers}

Figure \ref{fig:osteocad-arch} presents the formal architecture of OsteoCAD, comprising three layers: Client Services, AI Services, and Infrastructure. Each layer separates data management, AI orchestration, visualization, and inference, ensuring scalability, maintainability, and seamless integration with existing clinical workflows. Furthermore, these layers enable end-to-end management of the full healthcare data life cycle, from the ingestion of raw DICOM files to their processing via AI services.

\textbf{Client Services}. This layer provides high-level functionality to end users (e.g., researchers or physicians) through a suite of web-based services. OsteoCAD currently offers tools for visualizing medical images, loading and creating datasets, saving and exporting data, configuring training and inference models, and visualizing performance metrics. The services in this layer can easily be extended to meet an organization's specific requirements.

\textbf{OsteoCAD AI Services}. Training and inference capabilities are provided to the client independently of their technical expertise. It includes a Data Ingestion module that preprocesses inputs by removing artifacts from medical images. This module handles images in DICOM and NIfTI formats, routing them to the next components: the Training Orchestrator and the Inference Engine. The Training Orchestrator enables users to configure different DL models using pre-annotated data, while the Inference Engine allows users to run predictions on their data using trained models. The final module in this layer is Data Export, which handles the outputs of both training and inference, including model weights, NIfTI volumes, and segmentation labels.

\textbf{Infrastructure}. Foundational base of the application, provides resources for training and inference tasks execution, orchestrating both local and remote computing environments depending on client needs. Local hardware processes and stores data that remain within the organization. A Data Manager module ingests datasets into OsteoCAD and preserves trained models in a Registry; it also routes data to compute nodes for model training and retrieves the resulting models and outputs. When local hardware is constrained (e.g., a scarcity of GPUs for training large models), the organization can connect to remote environments via a Security Controller. This component establishes secure SSH tunnels for data exchange. Note that we are currently integrating more robust data exchange strategies into OsteoCAD to further enhance privacy and security, building on methods we have previously tested~\cite{Torres-Charles2025Xook-Sec:Al.}. Remote resources operate as a Secure Sandbox equipped with an in-memory store, avoiding persistent local data storage; once the data has been used for a task, it is immediately deleted.

\subsection{Deployment Configurations}
\label{subsec:deployment-configs}

The OsteoCAD system can be configured to align with a hospital’s available resources, enabling seamless switching between local and remote execution modes. For example, a hospital may utilize a remote GPU cluster for computationally intensive training tasks while performing inference locally. Similarly, single-patient inference and visualization can be performed locally, whereas large-batch processing can be offloaded. This flexible configuration enables hospitals to balance operational time constraints with their available hardware.

To address the stringent privacy and security constraints inherent in healthcare data, OsteoCAD can be coupled with zero-trust data exchange policies when using remote resources. Crucially, before any data leaves the local hospital infrastructure, the Data Manager can be configured with a module to implement an automated de-identification protocol that removes Protected Health Information (PHI). Furthermore, all data in transit is secured via end-to-end encrypted SSH tunnels. On the remote end, the execution environment operates as an ephemeral, zero-trust sandbox. Decrypted data exists solely in volatile memory during model training or inference; no data is ever written to persistent storage, and the memory state is immediately purged upon task completion. 

\subsection{Data Pipeline and AI Integration}
\label{subsec:data-pipeline}

Currently, the OsteoCAD deep learning engine supports the nnU-Net framework, allowing full utilization of its training, ensembling, and post-processing capabilities. The resulting models and evaluation metrics are presented through the web client, enabling clinicians to select the most appropriate model for deployment (with the best-performing model provided as the default option).

Across the architecture, a centralized data pipeline manages the flow of medical data, beginning with the ingestion of raw DICOM files from clients. During training phases, this data is paired with annotations in well-known formats, such as CVAT 1.1. The pipeline automatically transforms this data into AI-ready NIfTI volumes through preprocessing steps, including artifact removal tailored for bone tumor imaging \cite{Rodriguez-Herrero2026AScans}. Finally, datasets are constructed with train/validation/test splits and fed into the model training module, which exports the resulting segmentations in standard formats ready for PACS integration.

\section{Case Study on the Detection of Tumors on Large Bone Images}
\label{sec:clinical-usecase}

We evaluated OsteoCAD through a clinical case study conducted in collaboration with the Luis Guillermo Ibarra Ibarra National Rehabilitation Center (INR-LGII) in Mexico. The study deployed a platform prototype with two main objectives: (1) creating a dataset where none previously existed and (2) developing a functional DL segmentation model for large-bone tumors, given that the institution specializes in the treatment of patients with these lesions. The entire workflow of the collaboration is detailed in Figure~\ref{fig:hitl-workflow} and explained below.

\subsection{Deployment testbed and dataset}

To overcome local resource limitations at INR-LGII, we deployed a secure, remote sandbox on GPU-equipped nodes, part of the C3 cluster at Universidad Carlos III de Madrid.

For this research, INR-LGII approved CT scans from \textbf{67 patients} with bone tumors (approximately $20{,}000$ images), 16 of which were initially annotated by an expert. The CT studies were shared exclusively for research purposes following approval by the institutional ethics board, and all scans were fully anonymized prior to sharing. To the best of our knowledge, no other large, publicly available bone tumor dataset with expert annotations currently exists. Therefore, the approved dataset will be maximally utilized, recognizing that manual annotation is a highly time-consuming task for clinicians. This limitation further underscores the major data scarcity challenge that continues to constrain the application of AI in medical imaging.

\subsection{Phase 1: Training pipeline}

The first phase of the workflow constructed in this case study consists of a three-step pipeline: dataset construction, remote training, and model selection and deployment.

In the dataset construction step, CT scans and CVAT-generated annotations from 16 patient cases were ingested into OsteoCAD to create the initial dataset structure, 20\% automatically held out as a test set. Although the initial cohort of labeled cases was small, DL-based data augmentation and the curated nnU-Net training pipeline were sufficient to bootstrap a usable dataset. This approach also respected clinicians’ limited time: in later iterations, experts refined automatic annotations instead of labeling from scratch. The Data Ingestion module converts both CT volumes and annotations into NIfTI format and organizes them according to the \textit{nnU-Net} expected directory structure. After this step, physicians can access two dataset variants: a raw dataset and a preprocessed version. 

Next, in the remote training step, the self-configuring \textit{nnU-Net v2} framework orchestrates the entire process, including dataset fingerprinting, data augmentation strategy selection, and U-Net configuration planning~\cite{Isensee2018NnU-Net:Segmentation}. 

For the raw dataset (512$\times$512 in-plane resolution), nnU-Net automatically generated three configurations: 2D, 3D full-resolution, and 3D low-resolution. After preprocessing (resulting in approximately 256$\times$256 resolution), only the 2D and 3D full-resolution configurations were applicable. It is worth noting that all aspects of architecture selection and training plan generation are automatically handled by the nnU-Net framework, based on the characteristics of the computing environment in which it is deployed.

Each configuration trains in a default 5-fold cross-validation scheme, with folds executed in parallel across five NVIDIA A40 GPUs on the C3 cluster (managed via SLURM job scheduling). The anonymized dataset was transferred via secure SSH tunnel to the remote sandbox, ensuring no patient data persisted post-training---all model artifacts (checkpoints, predictions) were returned to INR-LGII's local storage.

Final models for each configuration are obtained by ensembling the five folds' outputs, yielding five models total: raw dataset (2D, 3D low-resolution, 3D full-resolution); preprocessed dataset (2D, 3D full-resolution). This process was executed independently for both dataset variants.

In the final step of this pipeline, the inference model was selected. The best-performing model out of the five at this stage was the \textit{nnU-Net 3D Full-Resolution} model trained on the preprocessed dataset, achieving a Dice score of \textbf{$0.84 \pm 0.02$}. This model was subsequently packaged for local and batch inference. Note that a comprehensive comparison of alternative models lies beyond the scope of this paper, but it is transparently accessible from the OsteoCAD web interface as shown below.

\subsubsection{Phase 2: Inference, Refinement, and Expansion}

The second phase focused on applying the trained model to new cases for inference and iterative refinement. At the moment of writing this paper, we are evaluating the model using the remaining 51 patient studies in OsteoCAD. This evaluation will include participation by experts who will assess the model's segmentation outputs. For this purpose, OsteCAD provides interfaces that allow experts to view the results produced by the inference models.    

\begin{figure}[!b]
    \centering
    \includegraphics[width=\linewidth]{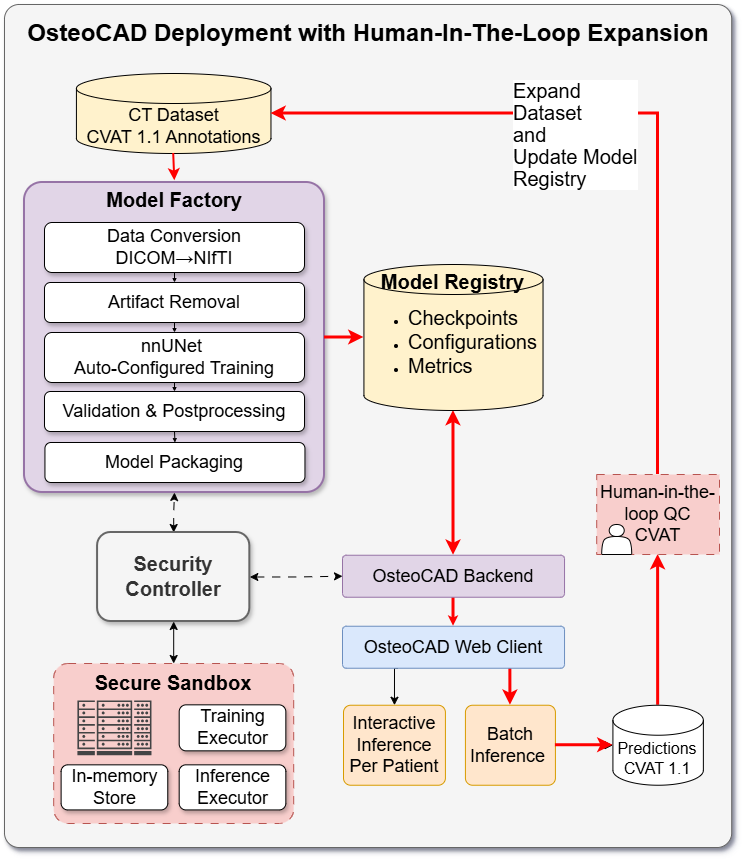}
    \caption{OsteoCAD workflow: From initial CVAT annotations to model factory (DICOM/NIfTI processing, nnU-Net training/QC, registry), enabling batch inference and human-in-the-loop expansion via CVAT corrections.}
    \label{fig:hitl-workflow}
\end{figure}

Figure~\ref{fig:hitl-workflow} illustrates the human-in-the-loop process used in this phase. The Model Factory handles DICOM preprocessing and inference, allowing experts to review and refine annotations via the OsteoCAD web client for continuous model retraining. These experts have access to pre-annotated CT scans via an OsteoCAD web client, which enables batch inference across multiple studies. Following this process, we can expand the original training dataset by including new studies. In our case study, this cycle---initial training on 16 patients, inference on the remaining 51, expert refinement, and retraining on all 67---enables sustainable dataset growth and improved model performance in a resource-constrained clinic without local GPU infrastructure.

\subsection{Inference examples}

\begin{figure}[!h]
    \centering
    \includegraphics[width=0.8\linewidth]{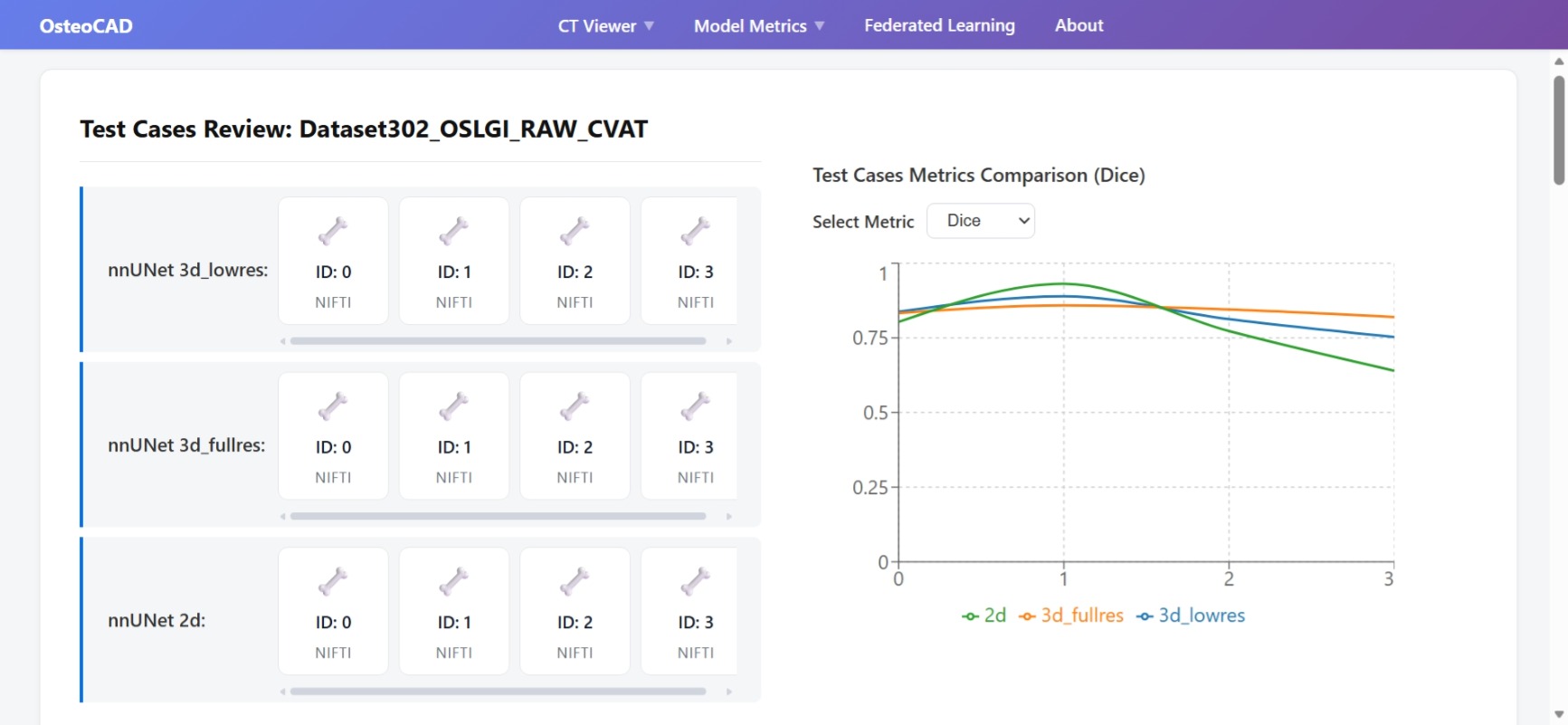}
    \hfill
    \includegraphics[width=0.8\linewidth]{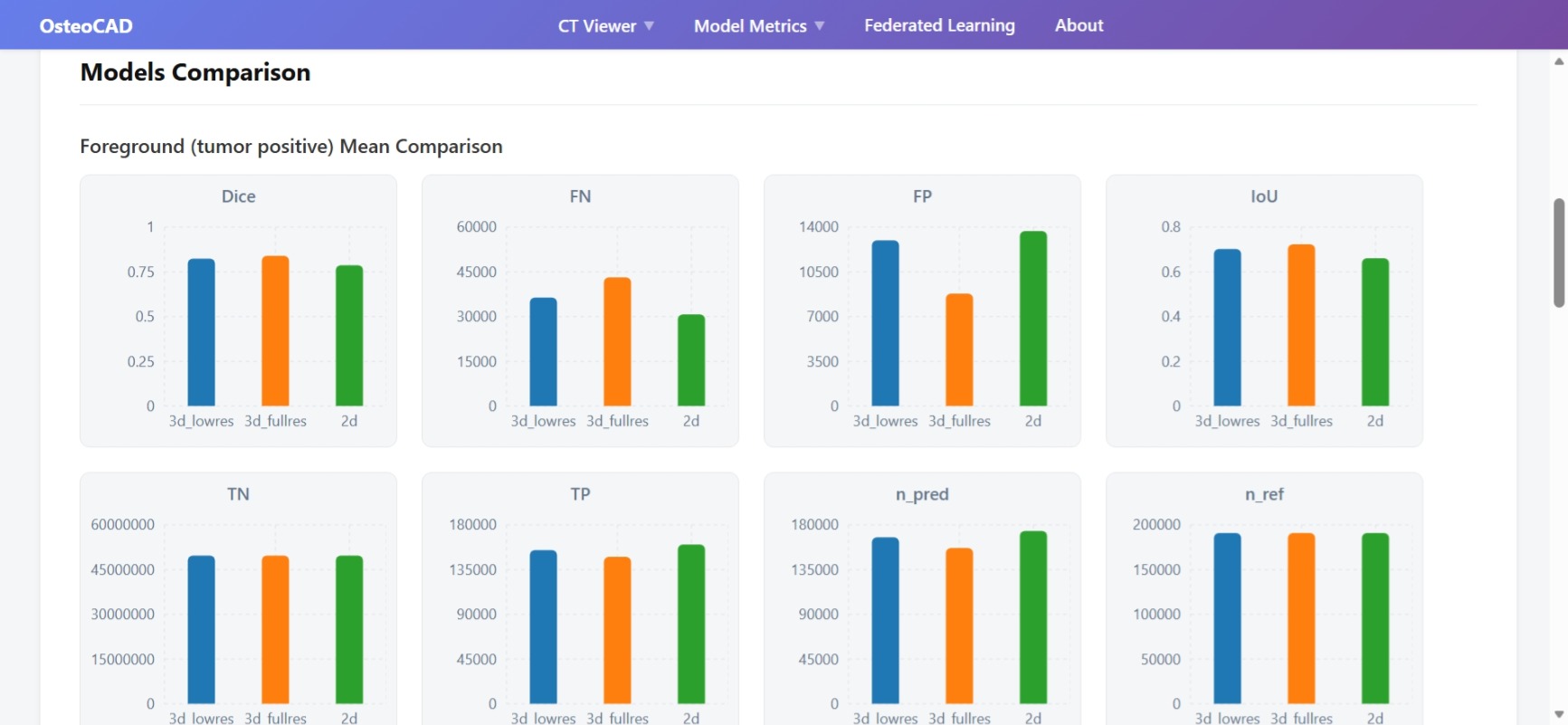}
    \caption{Model registry dashboard showing metrics (e.g. Dice), and redirection to test cases with ground truths and predictions for further inspection.}
    \label{fig:metrics-dashboard}
\end{figure}

OsteoCAD's web interface facilitates clinician interaction: model selection via metrics dashboard and 3D viewer for prediction/ground-truth comparison.
As shown in Figure~\ref{fig:metrics-dashboard}, the OsteoCAD dashboard aggregates nnU-Net metrics (Dice, Intersection Over Union, etc.) across final configurations and provides a direct link to the 3D viewer for test cases, facilitating comparison between predictions and ground truth. This helps researchers and managers to select a model without expertise.

\begin{figure*}
\centering
\begin{tikzpicture}
\node[anchor=center] (A) at (0,0)
    {\includegraphics[height=0.26\textheight]{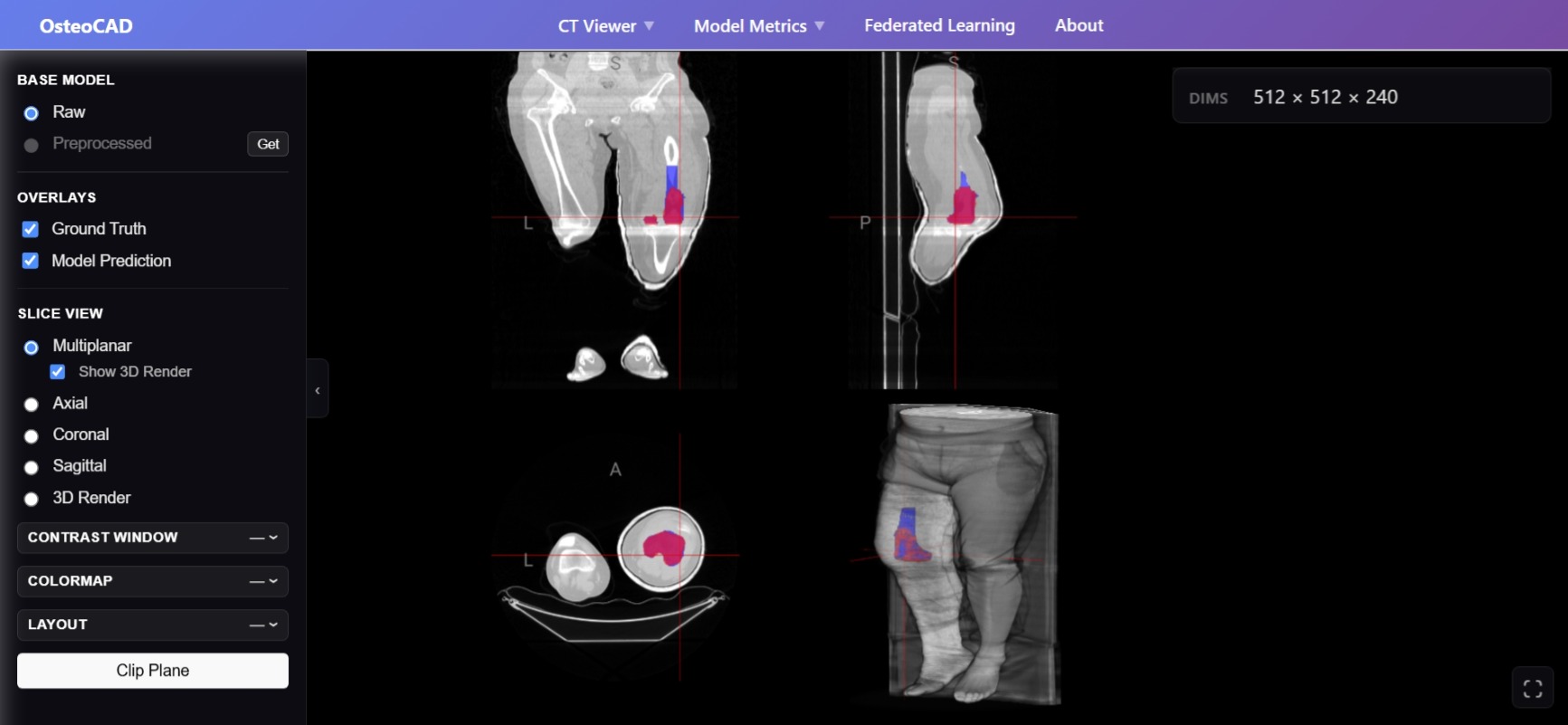}};
    \node[anchor=west, inner sep=1pt, fill=white] (B) at ($(A.east) - (4.5cm, 0)$)
        {\includegraphics[height=0.26\textheight]{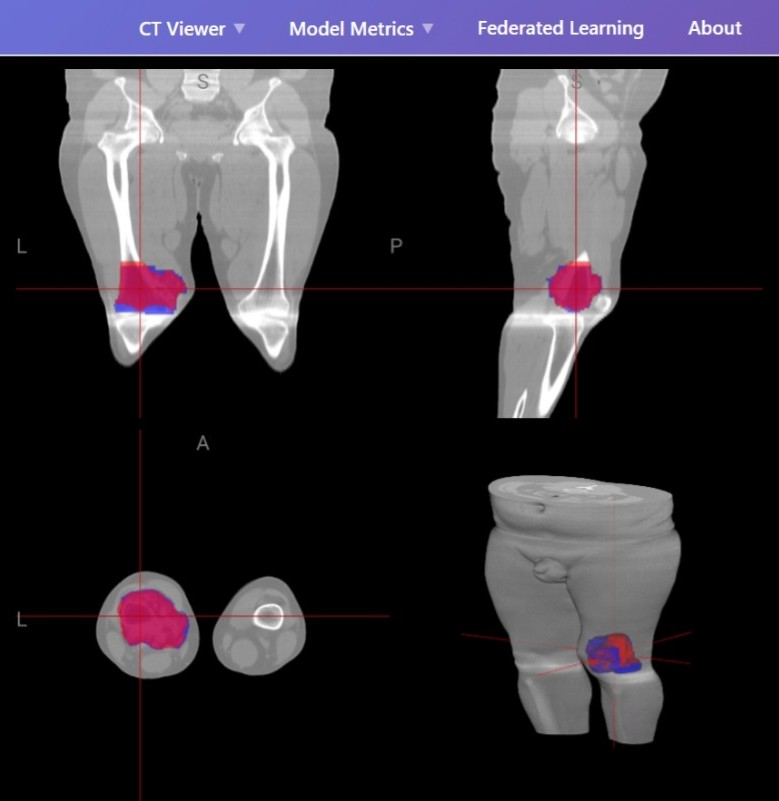}};
    \end{tikzpicture}
        \caption{3D medical viewer displaying original CT (grayscale), prediction (red), ground truth (blue) for bone tumor positive case. Left -- raw CT patient and model example, Right -- preprocessed CT and trained model example.}
\label{fig:3d-viewer}
\end{figure*}

Figure \ref{fig:3d-viewer} shows two real examples of the 3d viewer final result. Showcasing two test cases examples, one for the raw dataset (no artifact removal data trained model) and the preprocessed data (model trained over clean data) on it we see the cut views and 3D rendering model with overlays: tumor prediction (red) and ground truth (blue) for single-case validation. This UI facilitates patient inspection by directing professional attention to the potential region of interest. In the model development phase, it facilitates model inspection; all results shown in the viewer are exportable in standard formats for use in other applications or for direct integration with PACS or hospital existing software.

\section{Conclusions}
\label{sec:conclusions}

This work demonstrates that a modular framework like OsteoCAD can enable hospitals with minimal informatics support and no pre-existing annotated datasets to train, validate, and deploy state-of-the-art deep learning models in routine clinical workflows. By abstracting away infrastructure management, container orchestration, and GPU configuration, OsteoCAD lowers technical barriers for centers that are excluded from current AI adoption trends in medical imaging. Its emphasis on open standards, DICOM conventions, and exportable byproducts ensures data and model portability, avoiding vendor lock-in and enabling seamless integration with existing hospital tools such as 3D Slicer, where the final trained model can be used directly.

The pilot at INR-LGII demonstrates that maintaining continuity beyond proof of concept is essential for clinical integration. OsteoCAD supports this through a web-based interface, human-in-the-loop dataset scaling, and exportable artifacts (model weights, curated datasets, patient-level predictions) in standard formats. These enable clinicians to reuse segmentations in downstream workflows without disrupting established practices.

The rapid evolution of computing infrastructure and deep learning methods has made high-performing models for medical image analysis widely available. However, their routine use still depends on robust, maintainable infrastructure that many resource-constrained hospitals lack. At the same time, established institutions have demonstrated that AI can already be integrated into daily workflows when supported by dedicated engineering teams and specialized hardware. This asymmetry risks widening existing disparities in access to advanced diagnostic tools unless the underlying technology is deliberately adapted and democratized.

OsteoCAD addresses this asymmetry by providing a lightweight, extensible framework that adheres to modern software engineering and data governance principles. The pilot deployment confirms that it can stand up clinically useful segmentation models and build structured imaging datasets simultaneously, turning annotation effort into immediate clinical utility and future research assets.

Future work will advance OsteoCAD across three key areas. First, \textit{validation} will rely on in-clinic usability evaluation and continuous dataset expansion to ensure model robustness across diverse scanners and demographics. Second, \textit{implementation} will prioritize seamless PACS connectivity (e.g., XNAT, Orthanc) alongside enhanced enterprise security. Finally, core \textit{functionality} will be further expanded to enhance clinical trust through the integration of explainable AI techniques and automated patient report generation, consolidating model-derived insights such as tumor feature extraction. To improve generalization while preserving patient privacy and enabling scalable multi-center collaboration, federated learning approaches will be investigated.

\section*{Acknowledgment}
We acknowledge the C3 Cluster for computational resources, funded by the State Research Agency (AEI) through project ``Center for the Analysis and Modeling of Complex Systems in Engineering and Biomedicin'' (EQC2021-007184-P), supported by MICIU/AEI/10.13039/501100011033 and the EU NextGenerationEU/PRTR. This work was also partially funded by Universidad Carlos III de Madrid (PID 2025/00822/001; Proyectos Jóvenes PPIT2025 \& 09-PIN1-00005.4/2025) and by the R\&D project PID2022-138050NB-I00, funded by MICIU/AEI/10.13039/501100011033 ``FEDER: A way to make Europe.''


\bibliographystyle{IEEEtran}
\bibliography{references}

\end{document}